\documentclass[conference]{IEEEtran}
\usepackage{times}
\usepackage{epsfig}
\usepackage{graphicx}
\usepackage{amsmath}
\usepackage{amssymb}

\usepackage{algorithm}
\usepackage{algorithmicx}
\usepackage{algpseudocode}
\usepackage{subfigure}
\usepackage{multirow}

\begin{document}

\title{Pixel-level Reconstruction and Classification for Noisy Handwritten Bangla Characters}  






\author{\IEEEauthorblockN{Manohar Karki\IEEEauthorrefmark{1},
Qun Liu\IEEEauthorrefmark{1}, Robert DiBiano\IEEEauthorrefmark{2}\IEEEauthorrefmark{1},
Saikat Basu\IEEEauthorrefmark{1}, Supratik Mukhopadhyay\IEEEauthorrefmark{1}}

\IEEEauthorblockA{\IEEEauthorrefmark{1}Department of Computer Science,
Louisiana State University, Baton Rouge, LA, USA\\
}

\IEEEauthorblockA{\IEEEauthorrefmark{2}Ailectric LLC, Baton Rouge, LA, USA\\
\{mkarki6, qliu14, sbasu8\}@lsu.edu, robert@ailectric.com, supratik@csc.lsu.edu
}}

\maketitle

\begin{abstract}
Classification techniques for images of handwritten characters are susceptible to noise. Quadtrees can be an efficient representation for learning from sparse features. In this paper, we improve the effectiveness of probabilistic quadtrees by using a pixel level classifier to extract the character pixels and remove noise from handwritten character images. The pixel level denoiser (a deep belief network) uses the map responses obtained from a pretrained CNN as features for reconstructing the characters eliminating noise. We experimentally demonstrate the effectiveness of our approach by reconstructing and classifying a noisy version of handwritten Bangla Numeral and Basic Character datasets \cite{bhattacharya2009handwritten}, \cite{bhattacharya2012offline}.
\end{abstract}

\section{Introduction}
We build upon  recent developments in Deep Learning  to  develop techniques for efficient  representation of sparse features \cite{yosinski2014transferable, basu2015learning,deeptheory}. Sparse representations are usually compact representations of signals or features that have unnecessary default values \cite{huang2006sparse}\cite{boureau2008sparse}.  Quadtrees can be an efficient representation for learning from sparse features. Real world images, especially those of handwritten characters, are usually noisy.  Presence of noise can diminish the recognition power of classifiers \cite{noise1}\cite{noise2}. Efficient algorithms to remove noise can help in  classification.  We improve  the effectiveness of probabilistic quadtrees by using a pixel level classifier to extract the character pixels and remove noise from handwritten character images. 

\begin{figure}
   \begin{center}
     \includegraphics[width=.8\columnwidth]{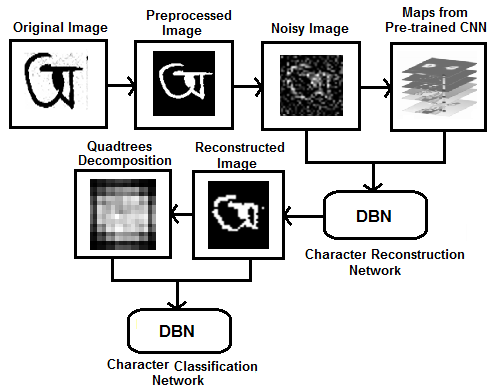}
    \caption{Architecture of our approach.}
    \label{fig:arch}
   \end{center} 
   \end{figure}

Figure \ref{fig:arch} shows the architecture of our approach. 
Our approach  uses a Convolutional Neural Network (CNN) pretrained on  the  ImageNet \cite{krizhevsky2012imagenet} collection  to extract features from images of handwritten characters.  These extracted features help  learn the shape of the characters and segment out those pixels that do not belong to the characters.  The information acquired  from  the pre-trained CNN helps   train  a Deep Belief Network (DBN) (called the reconstruction network) that reconstructs  the handwritten characters through  transfer learning \cite{yosinski2014transferable}, segmenting out noisy pixels. 

The  reconstruction network  takes as input a noisy  handwritten character image  and produces as output a  denoised binary version of it.
It segments out those pixels that do not belong   to the character.  A probabilistic quadtree is then used  to learn the sparse features  from the resulting denoised binary image obtained from  the reconstruction network  and  is used to  train a Character Classification Network (another DBN)   to classify the character images.

We experimentally demonstrate the effectiveness of our algorithm   at multiple resolutions using a recognition scheme similar to the one used in \cite{bhattacharya2009handwritten}  while improving on their performance. We also use the quadtree decomposition technique used  in \cite{basu2015learning} and improve upon it by using a saliency map to eliminate blocks that do not provide discriminating power. The saliency map   helps reduce the dimension of the training data  enabling  efficient learning. 
\subsection{Contributions}
This paper makes the following contributions:
\begin{itemize}
\item It develops an efficient  framework that removes noise from noisy images of handwritten characters
\item It improves  the effectiveness of probabilistic quadtrees by using a pixel level classifier to extract the character pixels  removing noise from handwritten character images.

\item It introduces a dataset 
comprising noisy Bangla basic characters and numerals with three different noise types. This dataset can serve as the basis and benchmark for future research in noisy Bangla character recognition.
\end{itemize}
     
       
\section{Related Work}
Data dimensionality reduction can   help  make   learning algorithms more efficient   while making the models simpler and  more robust \cite{haykin2004comprehensive}, \cite{hinton2006reducing} \cite{deepsat}, \cite{journal}. Quadtrees have been used previously to compress images \cite{markas1992quad} and represent spatial data \cite{aref1993decomposing}. In  \cite{ basu2015learning}, the authors use probabilistic quadtrees to represent  character images and classify them using a deep belief network (DBN). 



Map responses from intermediate layers of a Convolutional Neural Network have been used as features  in \cite{csf} to segment images.
  
  Chain code histogram features are used to discriminate classes in a multi stage approach in \cite{bhattacharya2009handwritten} where rejected images are classified again at higher resolutions.  We follow a similar approach of using higher resolutions when  classifying the  Bangla Numeral Dataset \cite{bhattacharya2009handwritten}. In \cite{bhattacharya2012offline} the authors   classify handwritten Bangla Basic Characters that have 50 classes. They use a two stage approach where they employ chaincode and gradient based features. The initial  stage uses a modified quadratic discriminant function (MQDF) based classifier  which is followed by  a Multi-layer Perception based classifier that helps to improve recognition on confused classes. 

  On the Noisy Bangla Datasets that we focus on, in \cite{basu2015learning} the authors have used probablistic quadtrees to learn sparse representations of handwritten character images and have used a two layer DBN for the classification.
  
  For handwritten character images containing significant amounts of  noise, the  sparse representations learnt  using quadtrees in \cite{basu2015learning}  are not efficient.  In \cite{basu2015learning}, the authors also introduce the  handwritten Bangla Numeral Dataset which includes the 10 Bangla Numeric Characters and  three types of noise: white gaussian noise, motion blur,  and reduced contrast. We inject these  three types of noises into the Bangla Basic Character dataset  consisting of 50 classes resulting in the Noisy Bangla Basic Character Dataset. In \cite{basu2015learning},  once a block has been chosen for decomposition  in any one image based on the homogeneity criterion, that block is identically decomposed for every other image. We  use a saliency map  instead which improves the representation and  we also use another DBN for character reconstruction  removing noise.
  \begin{figure}
   \begin{center}
     \includegraphics[width=.8\columnwidth]{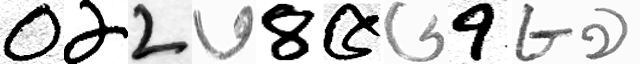}
    \caption{Original Images Bangla Numeral Characters. One image for each Bangla Numeral. \cite{bhattacharya2009handwritten}}
    \label{fig:orignum_train}
   \end{center} 
   \end{figure}
      
     \begin{figure}
   \begin{center}
     \includegraphics[width=.8\columnwidth]{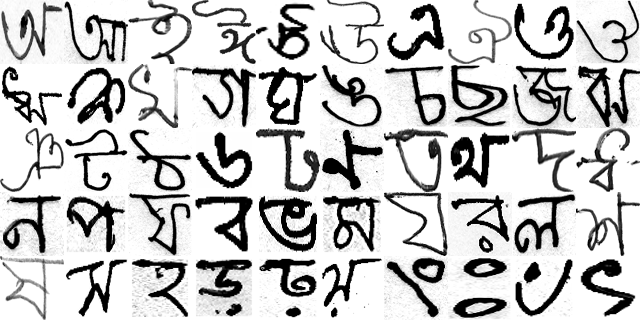}
    \caption{Original Images Bangla Basic Characters. One image each of the 50 Bangla Basic Characters.\cite{bhattacharya2012offline}}
    \label{fig:origbasic_train}
   \end{center} 
   \end{figure}

\section{Preprocessing Data}
The datasets that we evaluate our algorithm on consist of two types of handwritten  Bangla characters: a) Bangla Basic Characters b) Bangla Numeric Characters. There are 50 classes in the first type  and 10  in the second. 

\subsection{Standardize Raw Data}
The raw images are first processed using  the non-local means denoising algorithm \cite{buades2005non}. The resulting  images are bimodal in nature, with the pixels belonging to the background having value around one  and the rest, belonging to the character or noise, having lower values. The next step is to use Otsu's binarization scheme \cite{otsu1975threshold}  to threshold the images to binary. In the binarized images, we set the values of the pixels in the background to 0 and  those in the foreground  to 255. Following the procedure described in \cite{basu2015learning}, we then find the largest connected component for each image  and center the image around that component. We pad  the images, to create square images, resulting in at least 10 pixel long  borders on all sides. To avoid large boundaries, we crop images that have too many background pixels surrounding the characters. Fig. \ref{fig:origbasic_train} shows one sample image each from the 50 different classes of the Bangla Basic Characters dataset and Fig. \ref{fig:orignum_train} shows the 10 different Bangla Numerals.
 \begin{figure}
   \begin{center}
     \subfigure[added white gaussian noise (awgn)]{\includegraphics[width=\columnwidth]{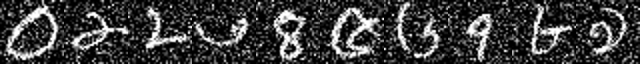}}
    \subfigure[motion blurred]{ \includegraphics[width=\columnwidth]{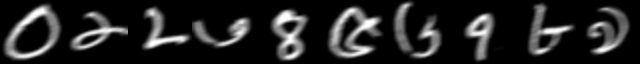}}
     \subfigure[decreased contrast and awgn]{  \includegraphics[width=\columnwidth]{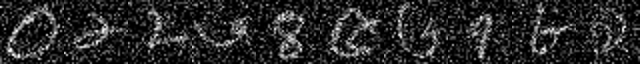}}
                  
    \caption{Noisy Version of the Bangla Numerals.}
    \label{fig:noisy}
   \end{center} 
   \end{figure}  

\subsection{Noisy Data Creation}
\begin{figure}
   \begin{center}
   \subfigure[added white gaussian noise (awgn)]{ \includegraphics[width=.78\columnwidth]{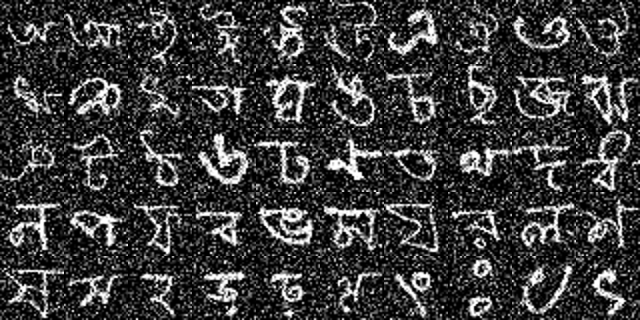}}
    \subfigure[decreased contrast and awgn]{ \includegraphics[width=.78\columnwidth]{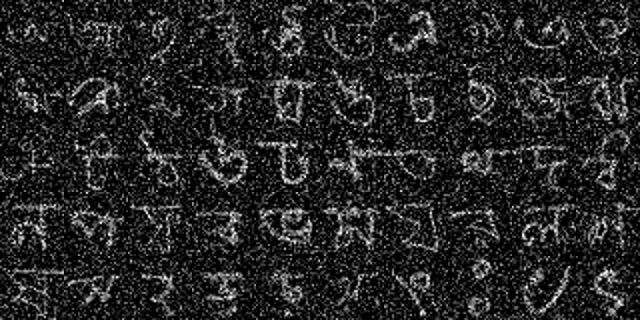}}
    \subfigure[motion blurred]{ \includegraphics[width=.78\columnwidth]{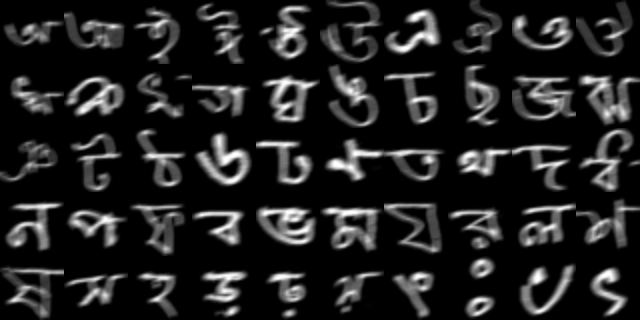}}
    \caption{Training Data for the Character Reconstruction Network. The noisy images of the the basic bangla characters}
    \label{fig:nt}
   \end{center} 
   \end{figure}

Three kinds of noisy  handwritten character image datasets  are created from the preprocessed dataset  by  adding three distinct types of noise. Similar to \cite{basu2015learning}, we create: a) the $\mathit{awgn}$ noisy dataset by adding  white Gaussian noise with a signal to noise ratio of 9.5 to the preprocessed dataset, b) the $\mathit{contrast}$ noisy dataset  by dividing  the intensity of the preprocessed  images by 2 and adding white Gaussian noise with a signal to noise ratio of 12,  and  (c) the $motion$ blurred noisy dataset by blurring the images with a linear motion of 5 and an angle of 15 degrees in the counterclockwise direction.  Fig. \ref{fig:noisy} shows the  samples from the noisy Bangla Numeral dataset for  each of the three noise types and Fig. \ref{fig:nt} shows the  samples from the noisy Bangla Basic Characters dataset for each of the three noise types. The three noise types we added are commonly found   in images due to  camera movements, poor illumination, high temperature, and movement of objects  \cite{zhou2017classification}. Natural images taken from cameras, scanner etc. often contain such noises that need to be taken into consideration during classification tasks.

\subsection{Ground Truth for the Character Reconstruction Network}\label{sec:gt}	
\begin{figure}
   \begin{center}  
  \includegraphics[width=.78\columnwidth]{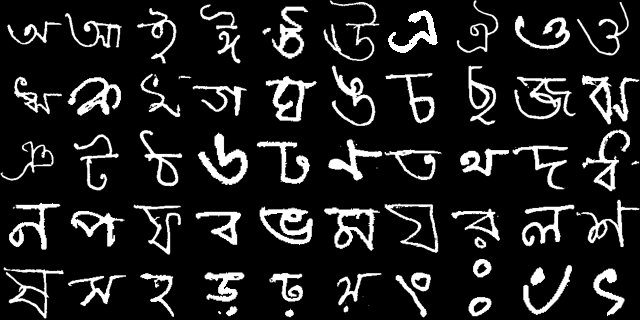}
\linebreak
    \caption{Ground truth sample images of the the basic bangla characters for the Character Reconstruction Network }
    \label{fig:gt}
   \end{center} 
   \end{figure}

As we do a pixel-level character reconstruction to clean the noisy images using the character reconstruction network, we use the binarized version of the images as labels for training the network. To keep the images binary after resampling, we use the nearest neighbor method. Each pixel in the ground truth can be grouped into one of the following two classes: a)  belonging to the background or b) belonging to the character/foreground. Fig. \ref{fig:gt} shows binary images used as ground truth for the Character Reconstruction Network.

\section{Deep Belief Networks}
A deep belief network is trained by first greedily pretraining  stacked Restricted Boltzmann Machines (RBMs) in an unsupervised way \cite{hinton2006reducing}.  A weight matrix for  an RBM is obtained by training it in an unsupervised manner. This weight matrix is then used to initialize the weights of the links connecting the first  and the second layers  of the DBN. The RBM then transforms the matrix of input feature vectors by calculating the mean activation of the hidden unit. The transformed data is then used to initialize the weights of the connections between the second and the third layer in a similar way. This process continues until the weights for the connections between the output layer and the one before it are initialized.  These weights are later fine-tuned using supervised training to change the learned representation to  a classifier. Because of the  unsupervised training in the beginning, DBNs are useful even with limited labeled data. Skip architectures are common in modern CNNs for pixel level classification for semantic segmentation \cite{long2015fully} that connect higher level layers to lower level ones. The use of DBN in our case is an alternative way to connect the map responses from different layers. There is  flexibility that comes with the network itself learning the combination of layer filters to use instead of explicit connections.

We use Deep Belief Networks at two stages of our approach. First, we use a DBN as a Character Reconstruction Network (CRN) to segment the pixels belonging to characters from the background   and then we also use a DBN as a Character Classification Network (CCN) to generate  the final classification using the feature vector representation provided by the probabilistic  quadtrees. The likelihood ($\mathcal{L}$) and loss ($J$) \cite{deeplearning.net} functions that we use in both the CRN and the CCN are given by:

\begin{equation}\label{eqn1}
\begin{split}
\mathcal{L} (\theta=\{M,B\},\mathcal{I})=\sum_{i=0}^{\left | \mathcal{I} \right |} log(P(Y=y_i|x_i,M,B))\\
J(\theta=\{M,B\},\mathcal{I})=-\mathcal{L}(\theta=\{M,B\},\mathcal{I}),
\end{split}
\end{equation}
where, $\mathcal{I}$ represents the dataset, and given an input $x_i$, the weights matrix $M$, and a bias vector $B$, $\mathcal{L}$ represents the likelihood that the input belongs to a class $y_i$. And, $J$ is the cost function,  the negative of  the log-likelihood of $x_i$ belonging to $y_i$ \cite{deeplearning.net}. For the CRN, this represents the likelihood that a pixel belongs to the character whereas for the CCN, this represents the likelihood that the whole image belongs to one of the character classes (either 10 or 50).

\section{Character Reconstruction Network}

The Character Reconstruction Network uses the map responses from the hidden layers of a previously trained CNN as features for pixel-wise classification. The CRN segments the pixels representing the characters from the rest of the pixels. It uses the map responses to the noisy images obtained when they are fed to the pretrained CNN as features.  Each pixel is treated as a single data point and the whole character image is reconstructed based on the classification of each pixel in that image. We do not know beforehand the type of noise present in an image. A single simple filter may not be enough to denoise images with unknown types of noise. 

\subsection{Transfer Learning} 
We  use transfer learning to extract information from the character images using the map responses from the pre-trained CNN as features to the CRN. The data that we feed in to either the CRN or the CCN  is not suitable for a convolutional network.  

 The ImageNet \cite{krizhevsky2012imagenet}, used to pretrain the CNN,    has 1000 object classes and more than a million images. It is trained mostly on objects, animals, scenes, and some geometric shapes.  Though we are dealing with a different type of data, it has been found that some of the higher level features learned by the layers of the convolutional layers are applicable in other types of images as well \cite{yosinski2014transferable}. Since the objective of the CRN is to use pixel-wise classification to segment the pixels belonging to the character from the rest and  the pre-trained CNN already takes into account the contextual information in the image, we use a DBN as the CRN. 

\subsection {Training the CRN} Our input to the Character Reconstruction Network is the preprocessed character images along with the maps extracted from the pre-trained CNN when these images are passed through it. The ground truth, as explained in  Section \ref{sec:gt}, is the binarized version of the original images. For the training of this network, all three types of noisy images, that includes  both Numeral and Basic Character images, are used.  We also train the CRN with  images without noise to make it more robust to images containing very little noise as well.  We only use images generated from  subset of the original dataset  reserved for training (and not the testing or validation set)  for the training, validation, and testing of this network. We sample 30 images each from all the three noisy versions of each of the two types of data: Bangla Numeral and Bangla Basic Character and 14 images each from the images without noise. We use the framework described in \cite{csf} to train the CRN where map responses from the pre-trained CNN are all rescaled to our input image size. Now, every  input image pixel  corresponds to a  pixel in each of the map responses. The values of the pixels  in the corresponding map responses are used as features that when  aligned together  form $hypercolumns$ \cite{hariharan2015hypercolumns}.  We train the Character Reconstruction Network by taking random samples from the pool of all available pixels along with their respective  features.
 \begin{figure}
   \begin{center}
       \subfigure[Noisy Samples of images (awgn) ]{\includegraphics[width=.78\columnwidth]{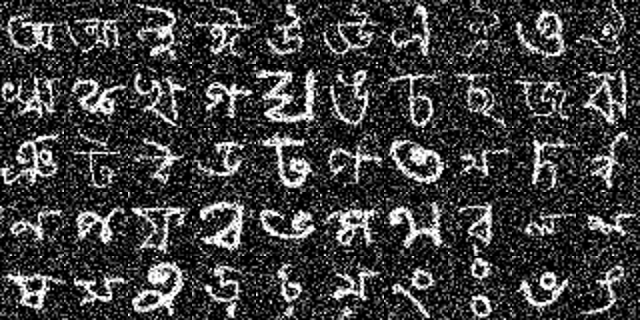}}
      \subfigure[Corresponding output]{ \includegraphics[width=.78\columnwidth]{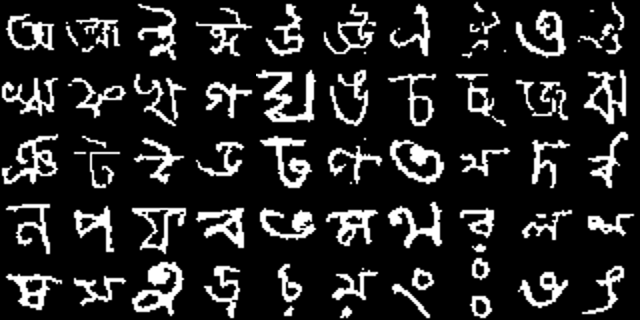}}
    \caption{Results from the Character Reconstruction Network}
    \label{fig:output_seg}
   \end{center} 
   \end{figure}

\section{Feature Representation using Quadtrees}
Decomposing an image window into maximal quadtree blocks has been used as an efficient way to represent sparse features \cite{aref1993decomposing}. We use this technique to represent the images. Considering that the use of quadtree decomposition is more effective in representing images with less noise, the previous step is done to denoise  noisy images. However, it is also beneficial to reduce dimensionality. Hence, we improve the sparse representation offered by probabilistic quadtrees \cite{basu2015learning} further by eliminating a) blocks   that have  been decomposed in only  a small fraction of samples and b) blocks that are present in almost all samples. This step  would make the dataset itself  less redundant.

\subsection{Saliency Map}

The homogeneity criterion described in \cite{basu2015learning} guides the process of reducing the character images into a vector of intensity values. Blocks containing textural details would be decomposed into smaller blocks. The quadtree representation is then converted to a linear vector by performing a depth first search (DFS). The features used consist of the averages  of the pixel values of the decomposed  blocks. While this approach reduces the number of features, when the data is noisy the quadtrees tend to be broken down into smaller blocks. Also, using this approach, whenever an image is broken into smaller blocks, all other images use this block in the final feature vector.

\begin{figure}
   \begin{center}
       \subfigure[Decomposition Map]{\includegraphics[width=.3\columnwidth]{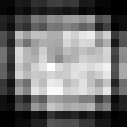}}    
      \subfigure[Saliency Mask]{ \includegraphics[width=.3\columnwidth]{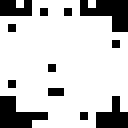}}
    \caption{Using saliency map to reduce dimensionality.}
    \label{maps}
   \end{center} 
   \end{figure}

 To mitigate this problem, we only use salient blocks that help in discrimination of characters. We use a saliency mask and prune smaller blocks that are not decomposed in $\mu$ percentage of the total number of training images or if they are contained in more than $\nu$ percentage of the samples. This technique decreases the number of features while not removing key blocks helpful to distinguish characters. The decomposition map on Fig. \ref{maps} (a) shows the normalized recurrence of the decomposed blocks for the entire training dataset for motion blurred  (noisy) Bangla Numeral images. The map shows that a lot of the blocks on the edges are hardly ever used, and some blocks in the middle are present in almost all of the images.  These blocks are not very likely to be useful in discriminating the characters. The saliency mask obtained in Fig. \ref{maps} (b) shows the blocks actually used. Table. \ref{numfeatures} shows the reduction of feature vectors using our technique compared to the probabilistic quadtree based approach.

 \subsection{Character Classification Network}

The Character Classification Network  is a DBN that uses the average pixel values of the different blocks  that have been decomposed in the  quadtree. We train the CCN to recognize the handwritten characters as belonging to individual character classes. In this case, the output is  a probability value that represents what character is present in the entire image instead of each pixel. The training data consists of the pixel values at all quadtree blocks. The labels are  derived from  the  labels of the images.

\begin{table}[]
\centering
\caption{Comparison of the number of features used}
\label{numfeatures}
\begin{tabular}{c|c|c|c}
Noise Type & Original & Ours & Basu et. al.\cite{basu2015learning} \\ \hline
awgn                & 1024              & \textbf{208}  & 244           \\
contrast            & 1024              & \textbf{203}  & 244           \\
motion              & 1024              & \textbf{196}  & 202          
\end{tabular}
\end{table}

\section{Experimental Results and Discussion}
We trained  both the CRN and the CCN  by first performing a pre-training step with persistent chain contrastive divergence (P-CD) and then fine tuned the network using   backpropagation. 
We used L1 and L2 norms for regularization and implemented dropout on the hidden layers. We ran our algorithms on  an Intel i7 six core server with TITAN X GPU and   used the  Theano deep learning library for the Deep Belief Networks \cite{2016arXiv160502688short}. We used $mu$ = 5\% and $nu$ = 95\% for our experiments. 

 On the Noisy Numeral Dataset, Table \ref{numeral_table} shows that we  significantly improve on  the results for each of the noise types in the Noisy Numeral dataset compared to \cite{basu2015learning}.  The use of the same network for reconstruction of all three types of noise hurt the performance on the $\mathit{motion}$  blurred images. We still make further improvements using the saliency mask. Observe that the classification accuracy increased even with a decrease in the number of features. We also compare our results against a multi layer neural network that makes use of Dropconnect for regularization \cite{wan2013regularization}. The approach has excellent results on the  MNIST classification with $<$ .5\% classification error \cite{wan2013regularization}. Interestingly the motion blurred images were very accurately classified by this approach. In Table \ref{dropconnect}, we show the classification error rate on the numeral and basic character images after they have been reconstructed using the CRN. The results show improvements especially on the images with added gaussian noise (awgn and contrast). In Table \ref{noisy_basic}, we compare our results on the Noisy Bangla Characters dataset with that obtained by  just using a traditional DBN using  raw pixels as features (similar to \cite{basu2015learning}) as well as those obtained using a multilayer neural network employing Dropconnect for regularization. We obtained  significant improvements in overall accuracy for each of the noise types using our approach.  We also studied the effect of various  CCN  architectures on the classification accuracy  on the noisy Bangla Basic Characters dataset in Table \ref{archbasic}. For the $\mathit{awgn}$ noise type, an architecture consisting of  two hidden layers with five hundred neurons per layer performed best; for the $\mathit{motion}$ noise type  an architecture consisting of  two hidden layers with three hundred neurons per layer performed best;   for  the $\mathit{contrast}$ noise type, the architectures consisting of  three hidden layers with five hundred  neurons per layer performed best  indicated in bold in the three columns of the Table \ref{archbasic}. 

\begin{table}[htbp]
\centering
\caption{Comparison of Error(\%) on the Noisy Bangla Numeral}
\label{numeral_table}
\begin{tabular}{c|c|c|c|c}
Noise    & Ours& Ours & Basu et. al.   &Dropconnect\\& (Saliency)& (w/o Saliency) &\cite{basu2015learning} &\cite{wan2013regularization} \\
\hline
awgn     & \textbf{4.54}& 4.92 & 8.66                                            & 8.82     \\

motion   & 4.96& 5.12 & 7.34                                             & \textbf{2.95}    \\

contrast & \textbf{7.15}& 7.4  & 12.69                                             & 14.21 

\end{tabular}
\end{table} 
\vspace*{-15pt}
\begin{table}[htbp]
\centering
\caption{ Classification Error(\%) using Dropconnect network after Noise Removal}
\label{dropconnect}
\begin{tabular}{c|c|c}
Noise    & Bangla Numeral & Bangla Basic Characters \\
\hline
awgn                  & 5.30             & 25.66 \\
motion         & 5.07               &24.96 \\
contrast               & 6.94           & 34.02 \\
\end{tabular}
\end{table}

\begin{table}[htbp]
\centering
\caption{Comparison of Error(\%) on Noisy Bangla Characters}
\label{noisy_basic}
\begin{tabular}{c|c|c|c|c}
Noise    & Ours & Ours &DBN & Dropconnect\\ & (Saliency)& (w/o Saliency) & (Raw Pixels) &\cite{wan2013regularization} \\
\hline
awgn                                                                  & \textbf{23.26}    & 29.36    & 42.69 &  38.86    \\
motion                                                                & 22.78     & 25.64      & 41.20  &  \textbf{16.41} \\
contrast                                                              & \textbf{30.34}     & 41.11       & 53.37  & 51.93  \\
\end{tabular}
\end{table}

\begin{table}[htbp]
\centering
\caption{Results on the Noisy Bangla Basic Characters Dataset with Various architecture}
\label{archbasic}
\begin{tabular}{c|ccc}
                                                                 & \multicolumn{3}{c}{Error (\%)}                                                             \\ \hline
\begin{tabular}[c]{@{}c@{}}Architecture\\ (Neurons)\end{tabular} & \multicolumn{1}{c|}{awgn}           & \multicolumn{1}{c|}{motion}         & contrast       \\ \hline
100 - 100                                                        & \multicolumn{1}{c|}{25.29}          & \multicolumn{1}{c|}{24.43}          & 33.63          \\
200 -200                                                         & \multicolumn{1}{c|}{23.57}          & \multicolumn{1}{c|}{23.26}          & 31.90          \\
300 - 300                                                        & \multicolumn{1}{c|}{23.47}          & \multicolumn{1}{c|}{\textbf{22.78}} & 31.16          \\
400 - 400                                                        & \multicolumn{1}{c|}{23.35}          & \multicolumn{1}{c|}{23.09}          & 30.71          \\
500 - 500                                                        & \multicolumn{1}{c|}{\textbf{23.26}} & \multicolumn{1}{c|}{23.35}          & 30.82          \\
1000 -1000                                                       & \multicolumn{1}{c|}{23.28}          & \multicolumn{1}{c|}{23.03}          & 30.88          \\
100 - 100 - 100                                                  & \multicolumn{1}{c|}{26.81}          & \multicolumn{1}{c|}{26.03}          & 39.90          \\
300 - 300 - 300                                                  & \multicolumn{1}{c|}{24.08}          & \multicolumn{1}{c|}{23.49}          & 32.20          \\
500 - 500 - 500                                                  & \multicolumn{1}{c|}{23.28}          & \multicolumn{1}{c|}{23.24}          & \textbf{30.34} \\
1000 - 1000 - 1000                                               & \multicolumn{1}{c|}{23.49}          & \multicolumn{1}{c|}{23.27}          & 31.31         
\end{tabular}
\end{table}
Table \ref{results_mnist} compares the results on the n-MNIST dataset provided in \cite{basu2015learning}. We improve the recognition rates on the datasets with added white gaussian noise (awgn) and with reduced contrast and comparable results on  the motion blurred image. Our approach yields better results than using the method in \cite{wan2013regularization} in the same two types of noise. Our reconstruction algorithm was able to improve upon the results achieved from just noisy images using \cite{wan2013regularization} as well.

 In both n-MNIST and noisy Bangla Basic Characters, reduced contrast images had the worst recognition rates. The reconstruction network eroded many  of the pixels from the characters of the reduced contrast images. The random noise pixels had similar intensity values to the character pixels. Because of that edges were not as sharp as the other two types of noisy images. Our approach was not the best on motion blurred images. Motion blurred images were easily recognized by architectures that already work great on normal (non-noisy) characters.


\section{Conclusions}
We improved the efficiency of probabilistic quadtrees by using a pixel level classifier to reconstruct noisy handwritten character images by segmenting out the noisy pixels. The pixel level denoiser (a deep belief network)  uses  the map responses obtained from a pretrained CNN trained on Imagenet as features  for reconstructing  the  characters eliminating noise.

\begin{table}[htbp!]
\centering
\caption{Error (\%) on Noisy MNIST}
\label{results_mnist}
\begin{tabular}{c|c|c|c|c}

\begin{tabular}[c]{@{}c@{}}Noise\end{tabular} 
& \begin{tabular}[c]{@{}c@{}}Ours \\(Saliency)\end{tabular}        & \begin{tabular}[c]{@{}c@{}}Basu et. al.\\  {\cite{basu2015learning}}\end{tabular}  & \begin{tabular}[c]{@{}c@{}}Dropconnect\\   (Noisy) \cite{wan2013regularization}\end{tabular} &\begin{tabular}[c]{@{}c@{}}Dropconnect\\ (Reconstructed)\end{tabular} \\ \hline

awgn                                                     & {\textbf{2.38}}  & {9.93}		&	 {3.98} &{2.43}\\
motion                                                     & {2.80}  & {2.60}	&	 {\textbf{1.42}}&{2.80} \\
contrast                                                      & {\textbf{4.96}}  	& {7.84}	&	{6.76} & {5.07} \\
\end{tabular}
\end{table}


\bibliographystyle{IEEEtran}
\bibliography{icdarbib}

\end{document}